\documentclass[10pt,twocolumn,letterpaper]{article}

\usepackage{cvpr}
\usepackage{times}
\usepackage{epsfig}
\usepackage{graphicx}
\usepackage{amsmath}
\usepackage{amssymb}

\usepackage[breaklinks=true,bookmarks=false]{hyperref}
\usepackage{multirow}

\cvprfinalcopy 

\def\cvprPaperID{0409} 

\ifcvprfinal\fi
\begin{document}

\title{Improving Referring Expression Grounding \\with Cross-modal Attention-guided Erasing}

\author{Xihui Liu$^1$~~~~~~
Zihao Wang$^2$~~~~~~
Jing Shao$^2$~~~~~~
Xiaogang Wang$^1$~~~~~~
Hongsheng Li$^1$\\
$^1$The Chinese University of Hong Kong~~~~~~~~~~~~~~~$^2$ SenseTime Research \\
{\tt\small \{xihuiliu, xgwang, hsli\}@ee.cuhk.edu.hk}~~~~~~~~~~~~{\tt\small \{wangzihao, shaojing\}@sensetime.com}
}

\maketitle
\thispagestyle{empty}

\begin{abstract}
Referring expression grounding aims at locating certain objects or persons in an image with a referring expression, where the key challenge is to comprehend and align various types of information from visual and textual domain, such as visual attributes, location and interactions with surrounding regions. Although the attention mechanism has been successfully applied for cross-modal alignments, previous attention models focus on only the most dominant features of both modalities, and neglect the fact that there could be multiple comprehensive textual-visual correspondences between images and referring expressions. To tackle this issue, we design a novel cross-modal attention-guided erasing approach, where we discard the most dominant information from either textual or visual domains to generate difficult training samples online, and to drive the model to discover complementary textual-visual correspondences. Extensive experiments demonstrate the effectiveness of our proposed method, which achieves state-of-the-art performance on three referring expression grounding datasets.
\end{abstract}

\vspace{-10pt}
\section{Introduction}

The goal of  referring expression grounding~\cite{kazemzadeh2014referitgame,yu2016modeling,mao2016generation} is to locate objects or persons in an image referred by natural language descriptions. Although much progress has been made in bridging vision and language~\cite{faghri2017vse++,vinyals2015show,nguyen2018improved,yin2018zoom,gao2018question,chen2018improving,liu2018show}, grounding referring expressions remains challenging because it requires a comprehensive understanding of complex language semantics and various types of visual information, such as objects, attributes, and relationships between regions. 

Referring expression grounding is naturally formulated as an object retrieval task, where we retrieve a region that best matches the referring expression from a set of region proposals. Generally, it is difficult to trivially associate phrases and image regions in the embedding space where features are separately extracted from each modality (\ie, vision and language). Previous methods~\cite{yu2018mattnet,hu2017modeling} proposed modular networks to handle expressions with different types of information. 
Another line of research explored attention mechanism, which mines crucial cues of both modalities~\cite{yu2018mattnet,deng2018visual,zhuang2018parallel}. 
By concentrating on the most important aspects in both modalities, the model with attention mechanism is able to learn better correspondences between words/phrases and visual regions, thus benefits the alignment between vision and language. 

\begin{figure}[t]
\centering
\includegraphics[width=1\linewidth]{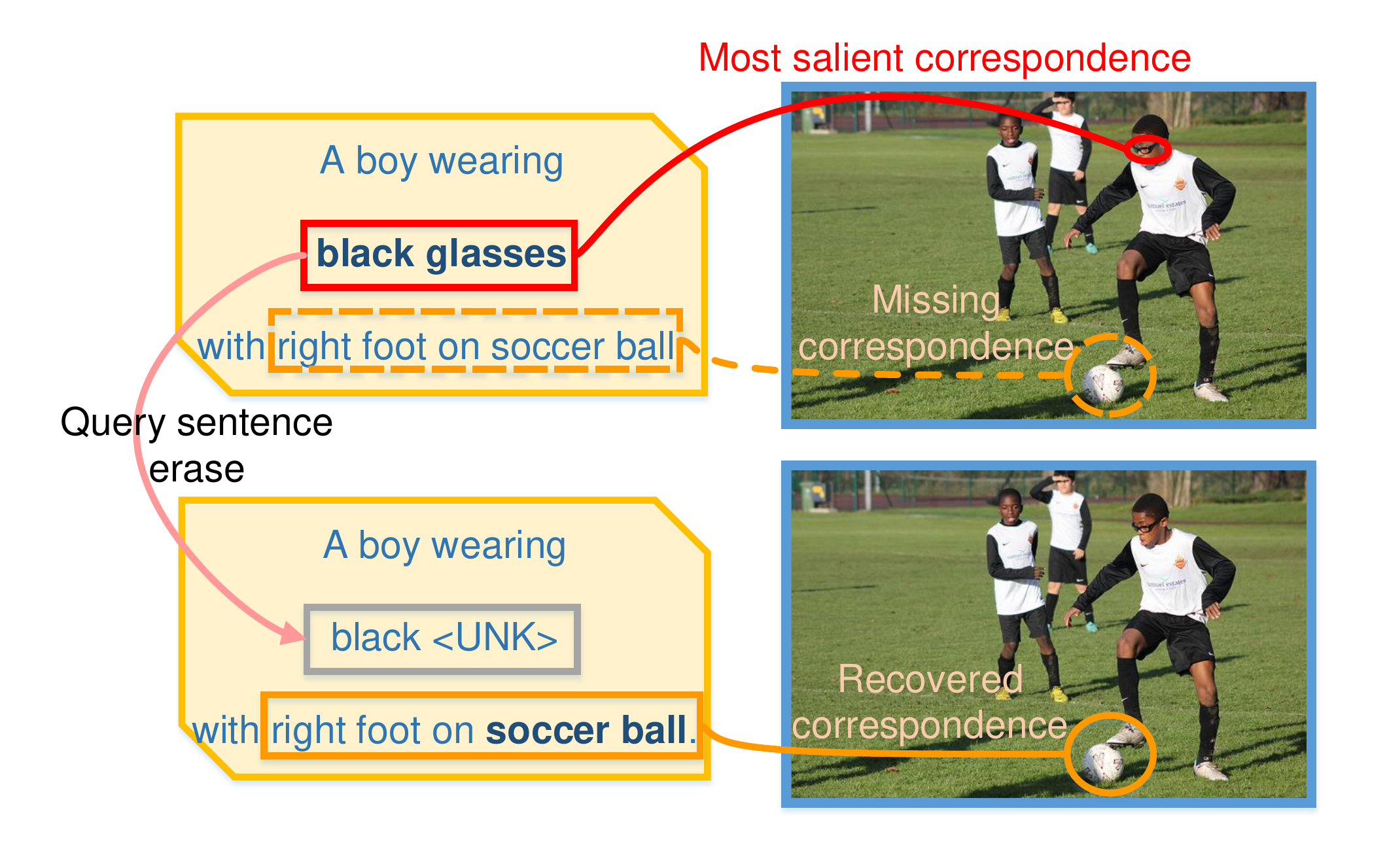}
\caption{Query sentence erasing as an example of our cross-modal attention-guided erasing. The first row shows the original query-region pair, and the second row shows the pair with erased query.}
\label{fig:intro}
\vspace{-10pt}
\end{figure}

However, a common problem of deep neural networks is that it tends to capture only the most discriminative information to satisfy the training constraints, ignoring other rich complementary information~\cite{zhang2018adversarial,wei2017object}. This issue becomes more severe when considering attention models for referring expression grounding.
By attending to both the referring expression and the image, the attention model is inclined to capturing the most dominant alignment between the two modalities, while neglecting other possible cross-modal correspondences. 
A referring expression usually describe an object from more than one perspectives, such as visual attributes, actions, and interactions with context objects, which cannot be fully explored by concentrating on only the most significant phrase-region pair. For example, people describe the image in Fig.~\ref{fig:intro} as ``A boy wearing black glasses with right foot on soccer ball''. We observe that the model gives most attention on word ``glasses'', while ignoring other information like ``soccer ball''. As a result, the model can achieve a high matching score as long as it is able to recognize ``glasses'', and would fail to learn the visual features associated with the words ``soccer ball''. 
We argue that such limitations cause two problems: (1) it prevents the model from making full use of latent correspondences between training pairs. (2) A model trained in this way could overly rely on specific words or visual concepts and could be biased towards frequently observed evidences. Although some works on the recurrent or stacked attention~\cite{zhuang2018parallel,deng2018visual} perform multiple steps of attention to focus on multiple cues, they have no direct supervision on attention weights at each step and thus cannot guarantee that the models would learn complementary alignments rather than always focusing on similar information.

Inspired by previous works~\cite{singh2017hide,wei2017object} where they erase discovered regions to find complementary object regions, we design an innovative cross-modal erasing scheme to fully discover comprehensive latent correspondences between textual and visual semantics.
Our cross-modal erasing approach erases the most dominant visual or textual information with high attention weights to generate difficult training samples online, so as to drive the model to look for complementary evidences besides the most dominant ones. Our approach utilizes the erased images with original queries, or erased queries with original images to form hard training pairs, and does not increase inference complexity.
Furthermore, we take the interaction between image and referring expression into account, and use information from both self modality and the other modality as cues for selecting the most dominant information to erase. In particular, we leverage three types of erasing: (1) \textbf{Image-aware query sentence erasing}, where we use visual information as cues to obtain word-level attention weights, and replace the word with high attention weights with an ``unknown'' token. (2) \textbf{Sentence-aware subject region erasing}, where the spatial attention over subject region is derived based on both visual features and query information, and we erase the spatial features with the highest attention weights. (3) \textbf{Sentence-aware context object erasing}, where we erase a dominant context region, based on the sentence-aware object-level attention weights over context objects. 
Note that (2) and (3) are two complementary approaches for sentence-aware visual erasing. With training samples generated online by the \textit{erasing} operation, the model cannot access the \textit{most} dominant information, and is forced to further discover complementary textual-visual correspondences previously ignored.

To summarize, we introduce a novel cross-modal attention-guided erasing approach on both textual and visual domains, to encourage the model to discover comprehensive latent textual-visual alignments for referring expression grounding. To the best of our knowledge, this is the first work to consider erasing in both textual and visual domains to learn better cross-modal correspondences. To validate the effectiveness of our proposed approach, we conduct experiments on three referring expression datasets, and achieve state-of-the-art performance.

\section{Related Work}

\noindent\textbf{Referring expression grounding.}
Referring expression grounding, also known as referring expression comprehension, is often formulated as an object retrieval task~\cite{hu2016natural,plummer2017conditional}. 
\cite{yu2016modeling,nagaraja2016modeling,zhang2018grounding} explored context information in images, and~\cite{strub2018visual} proposed multi-step reasoning by multi-hop Feature-wise Linear Modulation. Hu \etal~\cite{hu2017modeling} proposed compositional modular networks, composed of a localization module and a relationship module, to identify subjects, objects and their relationships. Subsequent work by Yu \etal~\cite{yu2018mattnet} built MattNet, which decomposes cross-modal reasoning into subject, location and relationship modules, and utilizes language-based attention and visual attention to focus on relevant components. \cite{rohrbach2016grounding,mao2016generation,luo2017comprehension,yu2017joint,liu2017referring} considered referring expression generation and grounding as inverse tasks, by either using one task as a guidance to train another, or jointly training both tasks. 
Our work is built upon MattNet, and encourages the model to explore complementary cross-modal alignments by cross-modal erasing.

\vspace{1pt}
\noindent\textbf{Cross-modal Attention.}
Attention mechanism, which enables the model to select informative features, has been proven effective by previous works~\cite{xu2015show,lu2017knowing,chen2017sca,anderson2017bottom,yang2016stacked,nguyen2018improved,kim2018bilinear,nam2016dual,lee2018stacked,liu2017hydraplus}. 
In referring expression grounding, Deng \etal~\cite{deng2018visual} proposed A-ATT to circularly accumulate attention for images, queries, and objects. Zhuang \etal~\cite{zhuang2018parallel} proposed parallel attention network with recurrent attention to global visual content and object candidates. 
To prevent the attention models from over-concentrating on the most dominant correspondences, we propose attention-guided erasing which generates difficult training samples on-the-fly, to discover complementary cross-modal alignments.

\vspace{1pt}
\noindent\textbf{Adversarial erasing in visual Domain.}
Previous works has explored erasing image regions for object detection~\cite{wang2017fast}, person re-identification~\cite{huang2018adversarially}, weakly supervised detection~\cite{singh2017hide,hou2018self} and semantic segmentation~\cite{wei2017object}. Wang \etal~\cite{wang2017fast} proposed to train an adversarial network that generates training samples with occlusions and deformations for training robust detector. Wei \etal~\cite{wei2017object} and Zhang \etal~\cite{zhang2018adversarial} proposed adversarial erasing for weakly supervised detection and segmentation, which drives the network to discover new and complementary regions by erasing the currently mined regions. 

Different from previous works which only erase in visual domain, we take a step further towards cross-modal erasing in both images and sentences.
More importantly, our approach only erases to create new training samples in the training phase, and does not increase inference complexity. 

\section{Cross-modal Attention-guided Erasing}

Our cross-modal attention-guided erasing approach erases the most dominant information based on attention weights as importance indicators, to generate hard training samples, which drives the model to discover complementary evidences besides the most dominant ones. This approach is independent of the backbone architecture, and can be applied to any attention-based structures without introducing extra model parameters or inference complexity. 
In our experiments, we adopt the modular design of MattNet~\cite{yu2018mattnet} as our backbone, because of its capability to handle different types of information in referring expressions.


\subsection{Problem Formulation and Background}\label{subsec_background}

\begin{figure}[t]
\centering
\includegraphics[width=1\linewidth]{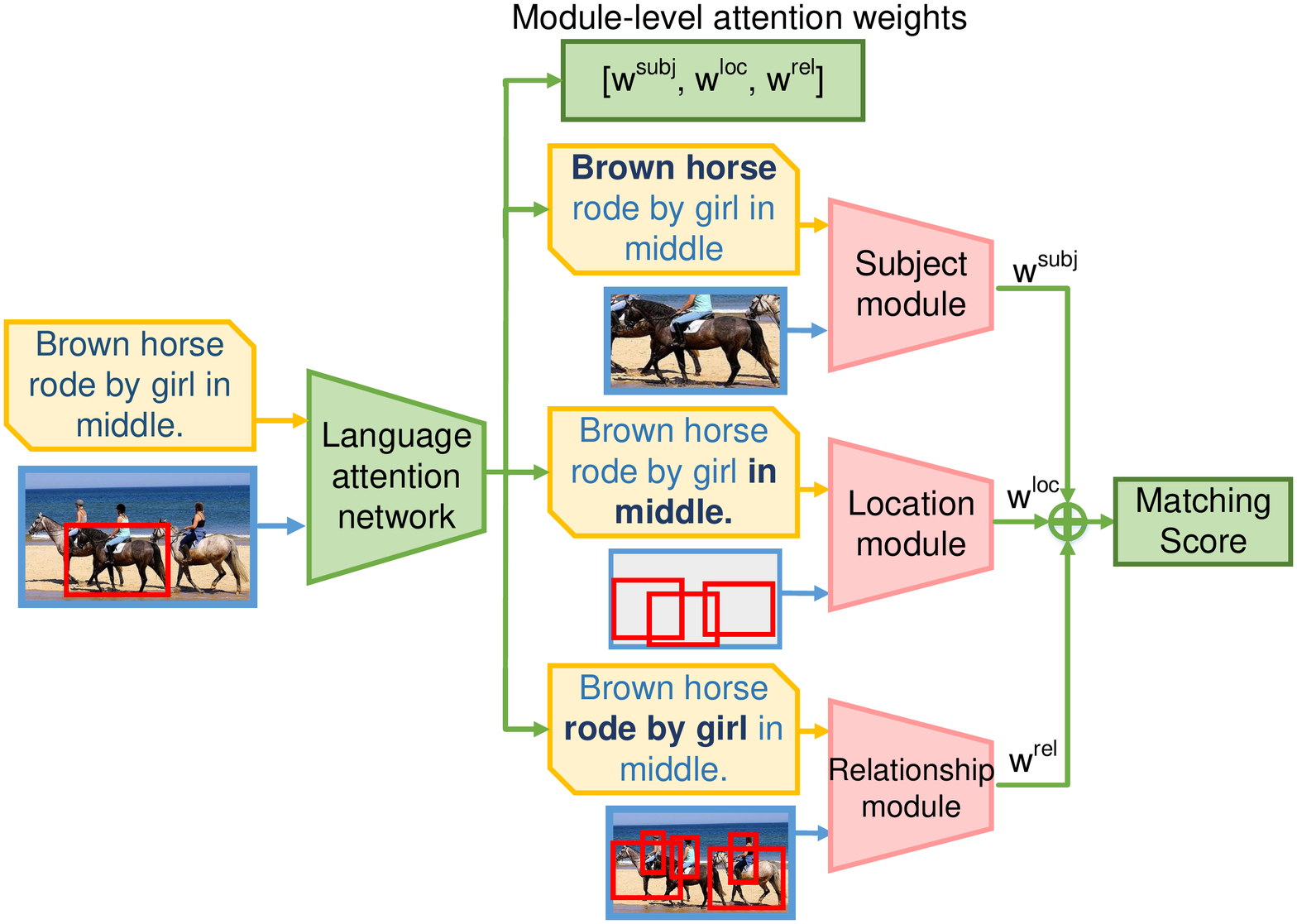}
\vspace{-20pt}
\caption{Illustration of our backbone model. The language attention network takes images and sentences as inputs, and outputs module-level attention weights and word-level attention weights for each module. The three visual modules calculate matching scores for subject, location and relationship, respectively. The final score is the weighted average of the three matching scores.}
\label{fig:baseline}
\vspace{-10pt}
\end{figure}

We formulate referring expression grounding as a retrieval problem: given an image $I$, a query sentence $Q$, and a set of region proposals $\mathcal{R}=\{R_i\}$ extracted from the image, we aim to compute a matching score between each region proposal $R_i$ and the query $Q$, and the proposal with the highest matching score is chosen as the target object. 
For each region proposal $R_i$, its regional visual features together with context object features are denoted as $O_i$.

In MattNet~\cite{yu2018mattnet}, there is a language attention network and three visual modules, namely subject module, location module and relationship module. 
The language attention network takes the query $Q$ as input, and outputs attention weights $\{w^{subj},w^{loc},w^{rel}\}$ and query embeddings for each module $[\mathbf{q}^{subj},\mathbf{q}^{loc},\mathbf{q}^{rel}]$.
Each module calculates a matching score by dot product between the corresponding query embedding and visual or location features.
The scores from three modules are fused according to the module-level attention weights $\{w^{subj},w^{loc},w^{rel}\}$. For positive candidate object and query pair $(O_i, Q_i)$ and negative pairs $(O_i, Q_j)$, $(O_j, Q_i)$, the ranking loss is minimized during training:
\vspace{-7pt}
\begin{align}\label{Lrank}
L_{rank} &= \sum_i ([m-s(O_i,Q_i)+s(O_i,Q_j)]_+ \nonumber \\
      &+ [m-s(O_i,Q_i)+s(O_j,Q_i)]_+),
\end{align}

\vspace{-10pt}

\noindent where $s(x, y)$ denotes the matching score between $x$ and $y$, $[x]_+=\max (x,0)$, and $m$ is the margin for ranking loss. 

We adopt the modular structure of MattNet~\cite{yu2018mattnet} and make some changes to the design of each module, which will be illustrated in Sec~\ref{subsec_sentence_erasing} to \ref{subsec_context_erasing}. The structure of our backbone is shown in Fig~\ref{fig:baseline}.


\subsection{Overview of Attention-guided Erasing}\label{subsec_overview}

By cross-modal erasing in both textual and visual domains to generate challenging training samples, we aim to discover complementary textual-visual alignments. 
(1) For \textbf{query sentence erasing}, we replace key words in the queries with the ``unknown'' token, and denote the erased referring expression as $Q^*$.
(2) For \textbf{visual erasing}, we first select which visual module to erase based on the modular attention weights. Specifically, we sample a module according to the distribution defined by the module-level attention weights $M^s \sim \text{Multinomial}(3, [w_{subj},w_{loc},w_{rel}])$, and perform erasing on the inputs of the sampled module. For subject module which processes visual information of candidate objects, we perform \textbf{subject region erasing} on feature maps. For location and relationship modules which encode location or visual features of multiple context regions, we apply \textbf{context object erasing} to discard features of a context object. The erased features by either subject region erasing or context object erasing is denoted as $O^*$. 

Given the erased query sentences or visual features, we replace the original samples with the erased ones in the loss function. Specifically, we force the erased visual features to match better with its corresponding queries than non-corresponding queries, and force the erased queries to match better with its corresponding visual features than non-corresponding ones, with the following erasing loss,
\vspace{-7pt}
\begin{align}\label{Lerase}
L_{erase} &= \sum_i ([m-s(O^*_i,Q_i)+s(O^*_i,Q_j)]_+ \nonumber \\
      &+ [m-s(O_i,Q^*_i)+s(O_j,Q^*_i)]_+).
\end{align}
\vspace{-17pt}

\noindent where the first term forces matching between the erased visual features and original queries, and the second term forces matching between the erased queries and original visual features.
We use a mixture of original and erased pairs in each mini-batch, and the overall loss is defined as,
\vspace{-5pt}
\begin{align}
L = L_{erase}+L_{rank}.
\end{align}
\vspace{-12pt}

In the following, we discuss how to perform the three types of cross-modal attention-guided erasing, respectively. 

\subsection{Image-aware Query Sentence Erasing}\label{subsec_sentence_erasing}

\begin{figure}[t]
\centering
\includegraphics[width=1\linewidth]{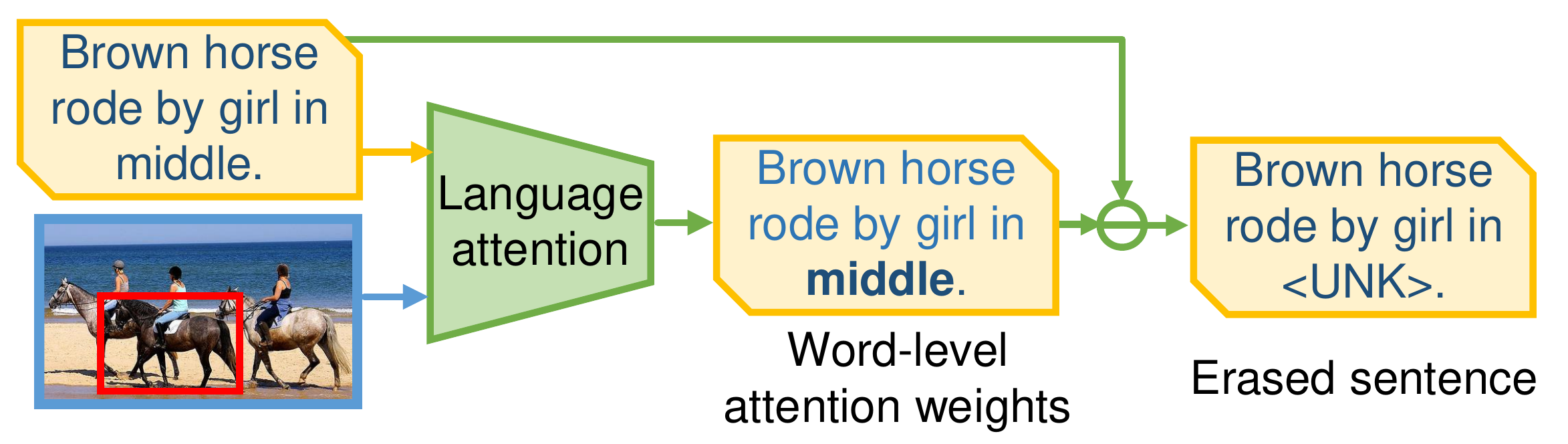}
\caption{Image-aware query sentence erasing.}
\label{fig:erase_lang}
\vspace{-10pt}
\end{figure}

People tend to describe a target object from multiple perspectives, but the model only focuses on the most dominant words, and neglects other words which may also imply rich alignments with visual information. Hence, we introduce erased queries into training to forbid the model from looking at only the most dominant word, so as to drive it to learn complementary textual-visual correspondences.

\vspace{3pt}
\noindent\textbf{Image-aware module-level and word-level attention.} 
Given the query sentence and the image, our first goal is to generate (1) attention weights for the three modules $\{w^{subj},w^{loc},w^{rel}\}$, and (2) three sets of word-level attention weights $\{\alpha_t^{subj}\}_{t=1}^T$, $\{\alpha_t^{loc}\}_{t=1}^T$, $\{\alpha_t^{rel}\}_{t=1}^T$ for three modules, where $T$ is the number of words in the sentence.

Generally, understanding a referring expression not only requires the textual information, but also needs the image content as a cue. Inspired by this intuition, we design an image-aware language attention network to estimate module-level and word-level attention weights.
Specifically, we encode the whole image $\mathbf{I}_0$ into a feature vector $\mathbf{e}_0$ with a convolutional neural network, and then feed the image feature vector and word embeddings $\{\mathbf{e}_t\}_{t=1}^T$ into the Long Short Term Memory Networks (LSTM). 
\vspace{-5pt}
\begin{equation}
\mathbf{e}_0 = \text{CNN}(\mathbf{I}_0), ~~~\mathbf{h}_t = \text{LSTM}(\mathbf{e}_t, \mathbf{h}_{t-1}).
\end{equation}
\vspace{-15pt}

\noindent We calculate the module-level and word-level attention weights based on the hidden states of the LSTM, and derive query embedding for each module accordingly,
\vspace{-7pt}
\begin{align}
w^m &= \frac{\exp(\mathbf{f}_m^T \mathbf{h}_T)}{\sum_{i\in \Omega} \exp(\mathbf{f}_i^T \mathbf{h}_T)},~~m \in \Omega, \\
\alpha_t^m &= \frac{\exp(\mathbf{g}_m^T \mathbf{h}_t)}{\sum_{i=1}^T \exp(\mathbf{g}_m^T \mathbf{h}_i)},~~~\mathbf{q}^m = \sum_{t=1}^T \alpha_t^m \mathbf{e}_t,
\end{align}
\vspace{-10pt}

\noindent where $\mathbf{f}_m$ and $\mathbf{g}_m$ are model parameters, $\Omega=\{\text{subj}, \text{loc}, \text{rel}\}$ represents the three modules, and $w^m$ denotes the model-level attention weights. $\alpha_t^m$ denotes the attention weight for word $t$ and module $m$, and $q^m$ is the query embedding for module $m$.

Our approach exploits visual cues to derive module-level and word-level attention weights, which is the key difference from previous works~\cite{yu2018mattnet,hu2017modeling} with only self-attention. 

\vspace{3pt}
\noindent\textbf{Attention-guided Query Erasing.}
Aiming to generate training samples by erasing the most important words in order to encourage the model to look for other evidences, we first calculate the overall significance of each word based on the module-level and word-level attention weights,
\vspace{-5pt}
\begin{equation}
\alpha_t = \sum_{m\in \Omega} w^m \alpha_t^m,
\end{equation}
\vspace{-10pt}

\noindent where $\{\alpha_t\}_{t=1}^T$ denotes the image-aware overall attention weight for each word, which acts as an indicator of word importance. We sample a word to erase based on the distribution defined by overall word-level significance, $W^s \sim \text{Multinomial}(T, [\alpha_1, ..., \alpha_T])$.

Next, we consider in what way shall we eliminate the influence of this word. The most straightforward way is to directly remove it from the query sentence, but the sentence grammar would be broken in this way. For example, if we directly remove the word ``chair'' from the sentence ``The gray office chair sitting behind a computer screen'', the overall semantic meaning would be distorted and the model might have difficulty understanding it. In order to eliminate the influence of the erased word while preserving the sentence structure, we replace the target word with an ``unknown'' token, as shown in Fig.~\ref{fig:erase_lang}. In this way we obtain the erased query $Q^*$, which discards the semantic meaning of the erased word, but causes no difficulty for the model to understand the remaining words. The erased query $Q_i^*$ and its original positive and negative image features $O_i$ and $O_j$ form new training sample pairs $(O_i, Q_i^*)$ and $(O_j, Q_i^*)$, and the we force textual-visual alignment between erased query sentences and original visual features by the ranking loss for erased query sentences (the second term in Eq.\eqref{Lerase}).

\subsection{Sentence-aware Subject Region Erasing}\label{subsec_pixel_erasing}

\begin{figure}[t]
\centering
\includegraphics[width=0.85\linewidth]{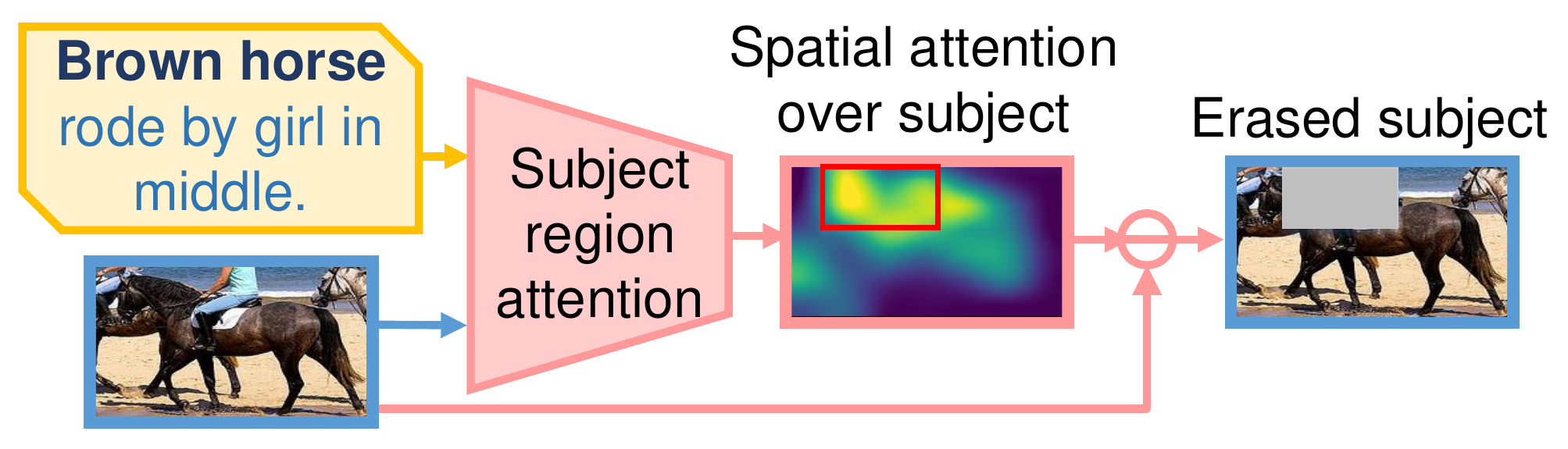}
\caption{Sentence-aware subject region erasing.}
\label{fig:erase_subj}
\vspace{-10pt}
\end{figure}

The subject module takes the feature map of a candidate region as input and outputs a feature vector. We create new training samples by erasing the most salient spatial features, to drive the model to discover complementary alignments.

\vspace{3pt}
\noindent\textbf{Sentence-aware spatial attention.}
We follow previous works on cross-modal visual attention~\cite{yu2018mattnet,yang2016stacked,deng2018visual}. For a candidate region with its spatial features $\{\mathbf{v_j}\}_{j=1}^J$, where $J$ is the number of spatial locations in the feature map, we concatenate the visual features at each location with the query embedding $\mathbf{q}^{subj}$ to calculate the spatial attention,
\vspace{-5pt}
\begin{align}
s_j &= \mathbf{w}_2^s\text{tanh}(\mathbf{W}_{1}^s[\mathbf{v}_j,\mathbf{q}^{subj}]+\mathbf{b}_1^s)+b_2^s, \\ 
\alpha_j^s &= \frac{\exp(s_j)}{\sum_{i=1}^J \exp(s_i)}, ~~~\mathbf{\tilde{v}}^{subj} = \sum_{j=1}^J \alpha_j^s \mathbf{v}_j,
\end{align}
\vspace{-7pt}

\noindent where $\mathbf{W}_{1}^s$, $\mathbf{w}_{2}^s$, $\mathbf{b}_1^s$, $b_2^s$ are model parameters, $s_j$ is the unnormalized attention, $\alpha_j$ is the normalized spatial attention weights, and $\mathbf{\tilde{v}}^{subj}$ is the aggregated subject features.

\vspace{3pt}
\noindent\textbf{Attention-guided subject region erasing.}
With conventional spatial attention, the model is inclined to focusing on only the most discriminative regions while neglecting other less salient regions. Such cases prevent the model from fully exploiting comprehensive textual-visual correspondences during training. So we erase salient features which are assigned greater attention weights to generate new training data, so as to drive the model to explore other spatial information and to learn complementary alignments.

In the feature map, spatially nearby features are correlated. Therefore, if we only erase features from separate locations, information of the erased features cannot be totally removed, since nearby pixels may also contain similar information. We therefore propose to erase a contiguous region of size $k \times k$ ($k=3$ in our experiments) from the input feature map. In this way, the model is forced to look elsewhere for other evidences. Particularly, we calculate the accumulated attention weights of all possible regions in the feature map by a $k \times k$ sliding window, and mask the region with the highest accumulated attention weights (See Fig.~\ref{fig:erase_subj} for illustration). The erased subject features together with original context object features are denoted as $O_i^*$. Similar to query sentence erasing, $O_i^*$ is paired with original query sentences to form positive training samples $(O_i^*, Q_i)$ and negative training samples $(O_i^*, Q_j)$, and the ranking loss for visual erasing (the first term in Eq.\eqref{Lerase}) is applied on the generated training sample pairs.

\subsection{Sentence-aware Context Object Erasing}\label{subsec_context_erasing}

\begin{figure}[t]
\centering
\includegraphics[width=0.9\linewidth]{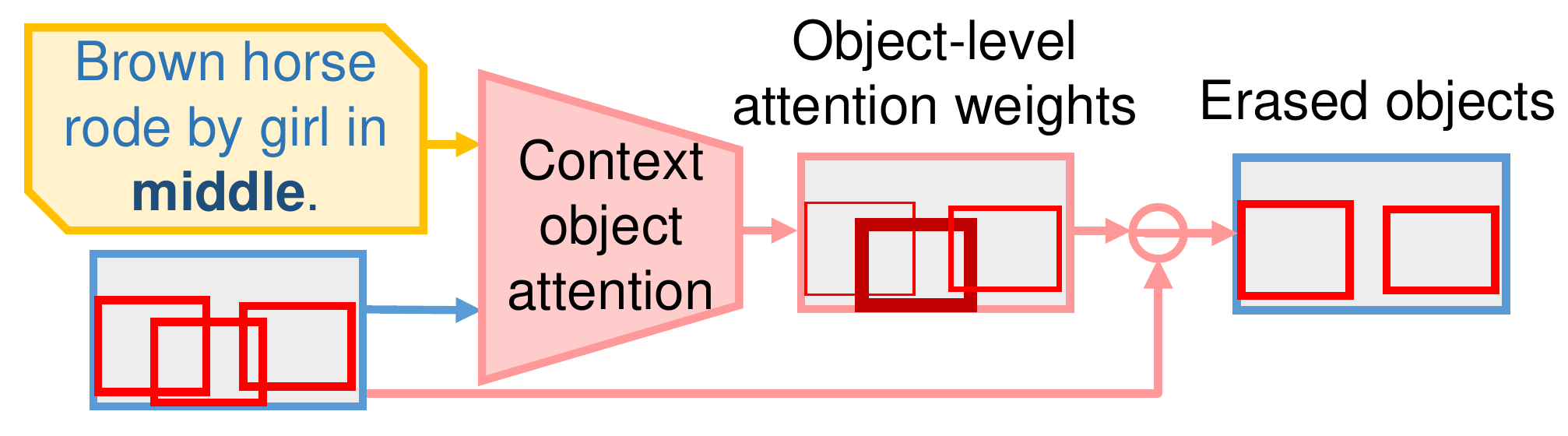}
\caption{Sentence-aware object erasing for location module.}
\label{fig:erase_loc}
\vspace{-10pt}
\end{figure}

In referring expression grounding, supporting information from context objects (\ie objects in the surrounding regions of the target object) is important to look for. For example, the expression ``The umbrella held by woman wearing a blue shirt'' requires an understanding of context region ``woman wearing a blue shirt'' and its relative location. 

\vspace{3pt}
\noindent\textbf{Sentence-aware attention over context objects.}
Sometimes multiple context regions are referred to in the sentence, \eg``White sofa near two red sofas''. 
So we formulate the location and relationship modules into a unified structure with sentence-aware attention, which considers multiple context objects, and attends to the most important ones.

For a set of context region features $\{\mathbf{c}^m_k\}_{k=1}^K$, where $m \in \{\text{loc}, \text{rel}\}$, and each $\mathbf{c}^m_k$ denotes the location or relationship feature of a context region proposal.\footnote{Details of context region selection and location and relationship feature extraction will be described in Sec~\ref{subsec_details}.} We derive object-level attention weights based on the concatenation of $\mathbf{c}^m_k$ and query embedding $\mathbf{q}^m$, and calculate the aggregated feature as the weighted sum of all object features,
\vspace{-8pt}
\begin{align}
s_k &= \mathbf{w}^m_2\text{tanh}(\mathbf{W}^{m}_1[\mathbf{c}^m_k,\mathbf{q}^{m}]+\mathbf{b}_1^m)+b_2^m, \\
a_k^m &= \frac{\exp(s_k)}{\sum_{i=1}^K \exp(s_i)}, ~~~\mathbf{\tilde{c}}^{m} = \sum_{k=1}^K a_k^m \mathbf{c}_k^m,
\end{align}
\vspace{-13pt}

\noindent where $\mathbf{W}_{1}^m$, $\mathbf{w}_{2}^m$, $\mathbf{b}_1^m$, $b_2^m$ are model parameters, $s_k$ is the unnormalized scores, $\alpha_k^m$ is the normalized object-level attention weights, and $\mathbf{\tilde{c}}^{m}$ is the aggregated module features.

Our unified attention structure for location and relationship modules is different from MattNet~\cite{yu2018mattnet}. 
In MattNet, the location module does not recognize different contributions of context regions, and the relationship module assumes only one context object contributes to recognizing the subject. 
In comparison, our model is able to deal with multiple context objects and attend to important ones, which is shown to be superior than MattNet in our experiments.

\vspace{3pt}
\noindent\textbf{Attention-guided context object erasing.}
Sometimes the model may find the target region with the evidence from a certain context object, and hence do not need to care about other information. So we leverage attention-guided context object erasing to discard a salient context object, and use the erased contexts to form training samples, to encourage the model to look for subject or other supporting regions.

\begin{figure}[t]
\centering
\includegraphics[width=0.95\linewidth]{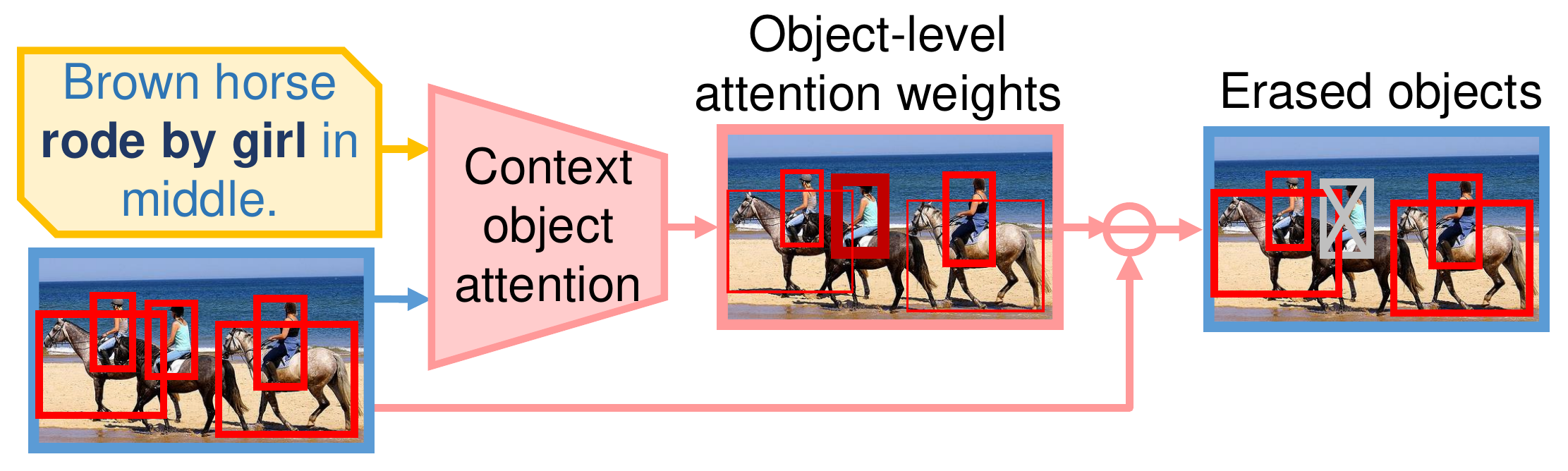}
\caption{Sentence-aware object erasing for relationship module.}
\label{fig:erase_rel}
\vspace{-10pt}
\end{figure}

For both location and relationship modules, we obtain object-level attention weights over all considered objects $\{\alpha_k^m\}_{k=1}^K$ by sentence-aware context object attention. We sample a context object according to the attention weights $C^s \sim \text{Multinomial}(K, [\alpha_1, ..., \alpha_K])$, and discard $C^s$ by replacing its features with zeros (see Fig.~\ref{fig:erase_loc} and Fig.~\ref{fig:erase_rel} for illustration). The erased context objects together with original subject features are denoted as $O_i^*$, which is paired with original query sentences to form positive training samples $(O_i^*, Q_i)$ and negative training samples $(O_i^*, Q_j)$, and the the ranking loss for visual erasing (the first term in Eq.\eqref{Lerase}) is applied on the generated training sample pairs. The erased samples will drive the model to look for other context regions or subject visual features, and to discover complementary textual-visual alignments. 

\subsection{Theoretical Analysis}\label{subsec_theoretical}

\vspace{3pt}
\noindent\textbf{Back-propagation Perspective.}
We derive the gradients of attention models, and reveal that it emphasizes the gradients of the most salient features while suppresses the gradients of unimportant features. Such a conclusion validates the necessity of our proposed attention-guided erasing.

Consider the visual modality with features $\{\mathbf{f}_i\}_{i=1}^m$ and attention weights $\{\alpha_i\}_{i=1}^m$, and the textual modality with features $\{\mathbf{g}_j\}_{j=1}^n$ and attention weights $\{\beta_j\}_{j=1}^n$. The aggregated features are $\mathbf{\tilde{f}}=\sum_{i=1}^m \alpha_i \mathbf{f}_i$ and $\mathbf{\tilde{g}}=\sum_{j=1}^n \beta_j \mathbf{g}_j$, respectively. We calculate the cross-modal similarity as, 
\vspace{-5pt}
\begin{align}
s &= \mathbf{\tilde{f}}^{\top}\mathbf{\tilde{g}} = \big(\sum_{i=1}^m \alpha_i \mathbf{f}_i\big)^{\top} \big(\sum_{j=1}^n \beta_j \mathbf{g}_j\big) = \sum_{i=1}^m \sum_{j=1}^n \alpha_i \beta_j \mathbf{f}_i^{\top} \mathbf{g}_j
\end{align}
\vspace{-15pt}

\noindent The gradient of $s$ with respect to $\alpha_i$, $\mathbf{f}_i$, $\beta_j$ and $\mathbf{g}_j$ are
\vspace{-5pt}
\begin{align}
\frac{\partial{s}}{\partial{\alpha_i}} &= \sum_{j=1}^n \beta_j \mathbf{f}_i^{\top} \mathbf{g}_j,~~~\frac{\partial{s}}{\partial{\mathbf{f}_i}} = \sum_{j=1}^n\alpha_i \beta_j \mathbf{g}_j, \\
\frac{\partial{s}}{\partial{\beta_j}} &= \sum_{i=1}^m \alpha_i \mathbf{f}_i^{\top} \mathbf{g}_j,~~~\frac{\partial{s}}{\partial{\mathbf{g}_j}} = \sum_{i=1}^m\alpha_i \beta_j \mathbf{f}_i.
\end{align}
\vspace{-10pt}

\noindent Suppose $s$ is the matching score between the corresponding candidate region and the query sentence, and receives a positive gradient during back-propagation. If $\mathbf{f}_i$ and $\mathbf{g}_j$ are close to each other and $\mathbf{f}_i^{\top} \mathbf{g}_j > 0$, the attention weights $\alpha_i$ and $\beta_j$ will receive positive gradients and be increased. On the contrary, if $\mathbf{f}_i^{\top} \mathbf{g}_j < 0$, both $\alpha_i$ and $\beta_j$ will be tuned down. As a result, attention mechanism automatically learns importance of features without direct supervision. 

On the other hand, if a word-region pair receives high attention $\alpha_i$ and $\beta_j$, the gradients with respect to $\mathbf{f}_i$ and $\mathbf{g}_j$ will be amplified, pushing $\mathbf{f}_i$ and $\mathbf{g}_j$ closer to each other to a large extent. While if $\alpha_i$ and $\beta_j$ are small, the gradients will be suppressed, only pushing $\mathbf{f}_i$ and $\mathbf{g}_j$ slightly closer to each other. As a result, the model would learn large attention and good alignments only for the best aligned features, and updates inefficiently for other cross-modal alignments with low attention weights. Inspired by this analysis, our approach erases the best aligned features, forcing the model to give high attention weights to complementary cross-modal alignments, and to update those features efficiently.

\vspace{3pt}
\noindent\textbf{Regularization Perspective.}
Our erasing mechanism can also be regarded as a regularization. The main difference from dropout~\cite{srivastava2014dropout} and dropblock~\cite{ghiasi2018dropblock} is that instead of randomly dropping features, we drop selectively. We erase salient information, as well as introducing randomness via sampling from the distributions defined by attention weights. The attention-guided erasing strategy is proven to be more effective than random erase in Sec.~\ref{subsec_ablation}.

\section{Experiments}
\vspace{-5pt}
\subsection{Implementation Details}\label{subsec_details}
\noindent\textbf{Visual feature representation.}
We follow MattNet~\cite{yu2018mattnet} for feature representation of subject, location and relationship modules. 
We use faster R-CNN~\cite{ren2015faster} with ResNet-101~\cite{he2016deep} as backbone to extract image features, subject features and context object features. Specifically, we feed the whole image into faster R-CNN and obtain the feature map before ROI pooling as the whole image feature (used in Sec.~\ref{subsec_sentence_erasing}). For each candidate object proposal, the $7 \times 7$ feature maps are extracted and fed into subject module (Sec.~\ref{subsec_pixel_erasing}). For the location module, we encode the location features as the relative location offsets and relative areas to the candidate object $\delta l_{ij}=\big[\frac{[\Delta x_{tl}]_{ij}}{w_i},\frac{[\Delta y_{tl}]_{ij}}{h_i},\frac{[\Delta x_{br}]_{ij}}{w_i},\frac{[\Delta y_{br}]_{ij}}{h_i}, \frac{w_j h_j}{w_i h_i}\big]$, as well as the position and relative area of the candidate object itself, \ie, $l_i=\big[\frac{x_{tl}}{W}, \frac{y_{tl}}{H}, \frac{x_{br}}{W}, \frac{y_{br}}{H}, \frac{w \cdot h}{W \cdot H}\big]$. Attention and erasing for location module in Sec.~\ref{subsec_context_erasing} is performed over the location features of up-to-five surrounding same-category objects plus the candidate object itself. For relationship module, we use the concatenation of the average-pooled visual feature from the region proposal and relative position offsets and relative areas $\delta l_{ij}=\big[\frac{[\Delta x_{tl}]_{ij}}{w_i},\frac{[\Delta y_{tl}]_{ij}}{h_i},\frac{[\Delta x_{br}]_{ij}}{w_i},\frac{[\Delta y_{br}]_{ij}}{h_i}, \frac{w_j h_j}{w_i h_i}\big]$ to represent relationship features of context objects. The attention and erasing on relationship module in Sec.~\ref{subsec_sentence_erasing} is performed over up-to-five surrounding objects.

\vspace{1pt}
\noindent\textbf{Training Strategy.} 
The faster R-CNN is trained on COCO training set, excluding samples from RefCOCO, RefCOCO+, and RefCOCOg's validation and test sets, and is fixed for extracting image and proposal features during training the grounding model. 
The model is trained with Adam optimizer~\cite{kingma2014adam} in two stages.
We first pretrain the model by only original training samples with ranking loss $L=L_{rank}$ to obtain reasonable attention models for erasing. Then, we perform online erasing, and train the model with both original samples and erased samples generated online, with the loss function $L=L_{rank}+L_{erase}$. 

\vspace{-3pt}
\subsection{Datasets and Evaluation Metrics}

We conduct experiments on three referring expression datasets: RefCOCO (UNC RefExp)~\cite{yu2016modeling}, RefCOCO+~\cite{yu2016modeling}, and RefCOCOg (Google RefExp)~\cite{mao2016generation}. 
For RefCOCOg, we follow the data split in~\cite{nagaraja2016modeling} to avoid the overlap of context information between different splits.

We adopt two settings for evaluation. In the first setting (denoted as ground-truth setting), the candidate regions are ground-truth bounding boxes, and a grounding is correct if the best-matching region is the same as the ground-truth. In the second setting (denoted as detection proposal setting), the model chooses the best-matching region from region proposals extracted by the object detection model, and a predicted region is correct if its intersection over union (IOU) with the ground-truth bounding box is greater than $0.5$. Since our work focuses on textual-visual correspondence and comprehension of cross-modal information, rather than detection performance, we report results under both settings, and conduct analysis and ablation study with the first setting.

\subsection{Results}
\vspace{-2pt}
\begin{table*}
\centering
\small
\begin{tabular}{c|c|ccc|ccc|ccc}
\hline
                                  &            & \multicolumn{3}{c|}{RefCOCO}         & \multicolumn{3}{c|}{RefCOCO+}                    & \multicolumn{3}{c}{RefCOCOg}              \\ \cline{3-11} 
                                &test setting & val        & testA    & testB    & val          & testA        & testB       & val$^*$ & val         & test           \\ \hline
MMI~\cite{mao2016generation}      &ground-truth & -       & 71.72     & 71.09    & -            & 58.42        & 51.23       & 62.14   & -           & -              \\
NegBag~\cite{nagaraja2016modeling}&ground-truth & 76.90   & 75.60     & 78.00    & -            & -            & -           & -       & -           & 68.40          \\
visdif+MMI~\cite{yu2016modeling}  &ground-truth & -       & 73.98     & 76.59    & -            & 59.17        & 55.62       & 64.02   & -           & -              \\
Luo~\etal~\cite{luo2017comprehension}&ground-truth & -       & 74.04     & 73.43    & -            & 60.26        & 55.03       & 65.36   & -           & -              \\
CMN~\cite{hu2017modeling}          &ground-truth & -       & -         &          & -            & -            & -           & 69.30   & -           & -              \\
Speaker/visdif~\cite{yu2016modeling}&ground-truth & 76.18   & 74.39     & 77.30    & 58.94        & 61.29        & 56.24       & 59.40   & -           & -              \\
S-L-R~\cite{yu2017joint}         &ground-truth & 79.56   & 78.95     & 80.22    & 62.26        & 64.60        & 59.62       & 72.63   & 71.65       & 71.92          \\
VC~\cite{zhang2018grounding}      &ground-truth & -       & 78.98     & 82.39    & -            & 62.56        & 62.90       & 73.98   & -           & -              \\
Attr~\cite{liu2017referring}    &ground-truth & -       & 78.05     & 78.07    & -            & 61.47        & 57.22       & 69.83   & -           & -              \\
Accu-Att~\cite{deng2018visual}    &ground-truth & 81.27   & 81.17     & 80.01    & 65.56        & 68.76        & 60.63       & 73.18   & -           & -              \\
PLAN~\cite{zhuang2018parallel}     &ground-truth & 81.67   & 80.81     & 81.32    & 64.18        & 66.31        & 61.46       & 69.47   & -           & -              \\
Multi-hop Film~\cite{strub2018visual}&ground-truth & 84.9    & 87.4      & 83.1     & 73.8         & \textbf{78.7}& 65.8        & 71.5    & -           & -              \\
MattNet~\cite{yu2018mattnet}    &ground-truth & 85.65   & 85.26     & 84.57    & 71.01        & 75.13        & 66.17       & -       & 78.10       & 78.12          \\ \hline
CM-Att                            &ground-truth & 86.23   & 86.57     & 85.36    & 72.36        & 74.64        & 67.07       & -       & 78.68       & 78.58          \\
CM-Att-Erase                    &ground-truth &\textbf{87.47}  &\textbf{88.12}   &\textbf{86.32}   &\textbf{73.74}   &77.58   &\textbf{68.85}   &-    &\textbf{80.23}&\textbf{80.37}\\\hline\hline
S-L-R~\cite{yu2017joint}          &det proposal& 69.48     & 73.71     & 64.96    & 55.71     & 60.74       & 48.80       & -    & 60.21      & 59.63      \\
Luo~\cite{luo2017comprehension}   &det proposal& -      & 67.94     & 55.18    & -         & 57.05       & 43.33       & 49.07   & -          & -          \\    
PLAN~\cite{zhuang2018parallel}     &det proposal& -        & 75.31     & 65.52    & -         & 61.34       & 50.86       & 58.03   & -          & -          \\ 
MattNet~\cite{yu2018mattnet}    &det proposal& 76.40    & 80.43     & 69.28    & 64.93     & 70.26        & 56.00     & -     & 66.67      & 67.01      \\ \hline
CM-Att                       &det proposal& 76.76   & 82.16     &  70.32   & 66.42        & 72.58        & 57.23       & -       & 67.32       & 67.55          \\
CM-Att-Erase                       &det proposal&\textbf{78.35}   &\textbf{83.14}   &\textbf{71.32}   &\textbf{68.09}   &\textbf{73.65}   &\textbf{58.03}   &-    &\textbf{67.99}&\textbf{68.67}\\\hline
\end{tabular}
\caption{Comparison with state-of-the-art referring expression grounding approaches on ground-truth regions and region proposals from detection model. For RefCOCO and RefCOCO+, testA is for grounding persons, and testB is for grounding objects.}
\label{tb:gt_results}
\vspace{-10pt}
\end{table*}
\vspace{-3pt}
\noindent\textbf{Quantitative results.}
We show results of referring expression grounding compared with previous works under the ground-truth setting and detection proposal setting in Table~\ref{tb:gt_results}. \textbf{CM-Att} denotes our model with cross-modal attention trained with only original training samples. \textbf{CM-Att-Erase} denotes our model with cross-modal attention trained with both original samples and erased samples generated by cross-modal attention-guided erasing. It is shown that the cross-modal attention model is already a strong baseline, and training with erased samples can further boost the performance. Our CM-Att-Erase model outperforms previous methods, without increasing inference complexity. It validates that with cross-modal erasing, the model is able to learn better textual-visual correspondences and is better at dealing with comprehensive grounding information.

\begin{figure}[t]
\centering
\includegraphics[width=1\linewidth]{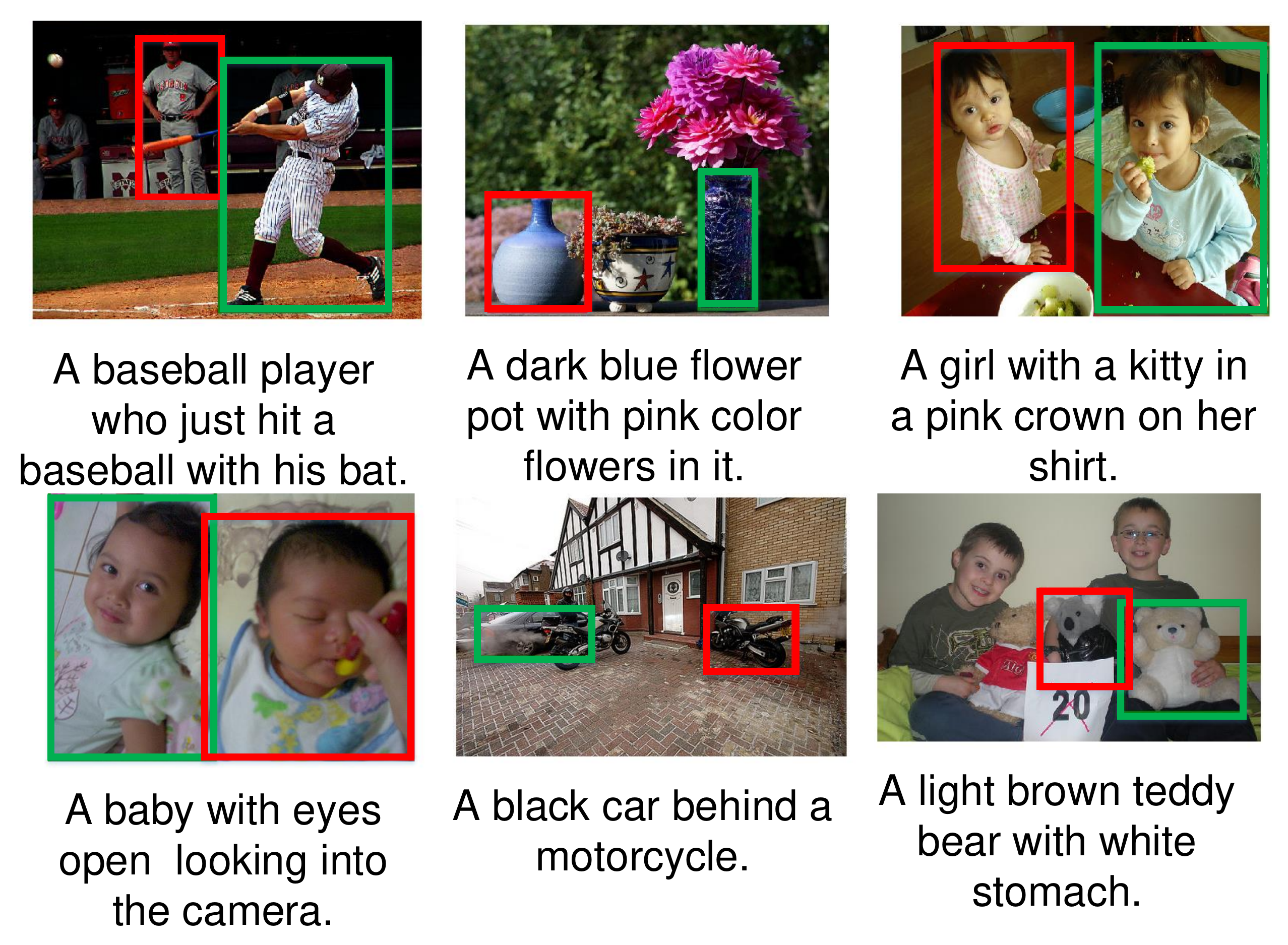}
\caption{Qualitative results. Red bounding box denotes the grounding results of the CM-Att model, and green bounding box denotes grounding results of the CM-Att-Erase model.}
\label{fig:results}
\vspace{-10pt}
\end{figure}

\vspace{1pt}
\noindent\textbf{Qualitative results.}
Fig.~\ref{fig:results} shows qualitative results of our CM-Att-Erase model, compared with the CM-Att model. It is shown that our CM-Att-Erase model is better at handling complex information from both domains, especially for situations where multiple cues should be considered in order to ground the referring expressions. Take the second image in the first row as an example, our erasing model comprehends not only visual features associated with ``dark blue flower pot'' but also relationship with context object ``pink flowers in it'', while the model without erasing does not perform well for those cases.

\subsection{Visualization of Attention and Erasing}

\begin{figure}[t]
\centering
\includegraphics[width=1\linewidth]{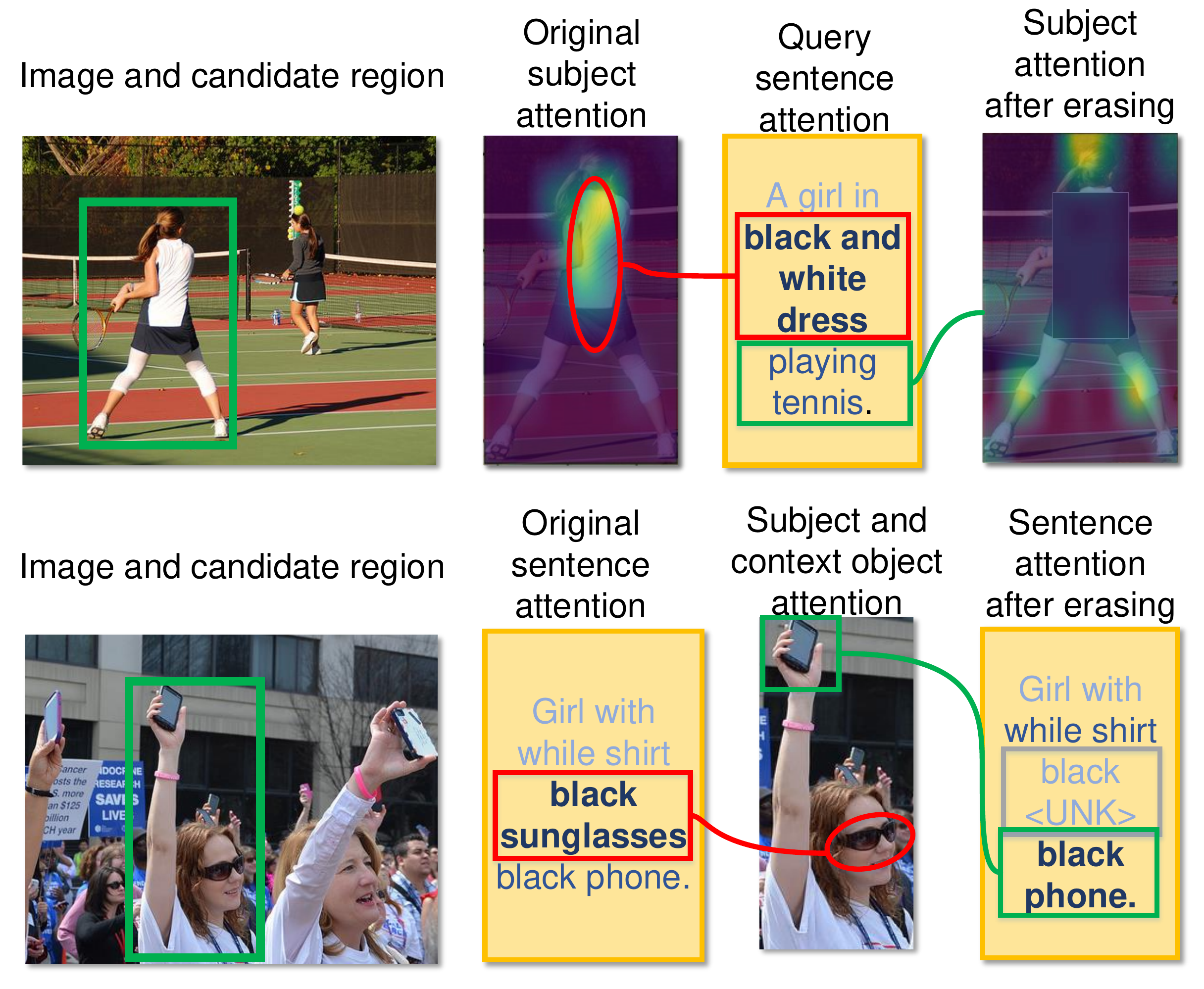}
\caption{Visualization of attention weights before and after erasing. The first line shows an example of subject region erasing, and the second line shows an example of query sentence erasing.}
\label{fig:visualize}
\vspace{-10pt}
\end{figure}

We visualize the attention weights and erasing process in Fig.~\ref{fig:visualize}. It is shown that in the first image, the subject module gives high attention weights to the region corresponding to ``black and white dress''. However after erasing this region, the subject module attends on the action of this girl, encouraging the model to learn the correspondence between ``playing tennis'' and its corresponding visual features. The second line shows an example of query sentence erasing. By erasing the word ``glasses'' to obtain a new erased query as training sample, the model is driven to look for other information in the image, and it successfully identifies the alignment between ``black phone'' and the corresponding context object in the image.

\subsection{Ablation Study}\label{subsec_ablation}
\begin{table}
\small
\begin{tabular}{cc|ccc}
\hline
                                                                                                                &                     & val       & test    \\ \hline
\multicolumn{2}{c|}{CM-Att-Erase (Our proposed approach)}                                                                       &   80.23   & 80.37   \\ \hline
\multicolumn{1}{c|}{\multirow{2}{*}{\begin{tabular}[c]{@{}c@{}}Erasing\\ methods\end{tabular}}}                 & Random              & 79.08     & 79.05   \\ \cline{2-2} 
\multicolumn{1}{c|}{}                                                                                           & Adversarial network to erase & 79.31     & 79.23   \\ \hline
\multicolumn{1}{c|}{\multirow{3}{*}{\begin{tabular}[c]{@{}c@{}}Effect of\\ Cross-modal\\ Erasing\end{tabular}}} & Self-erasing        & 79.27     & 79.22   \\ \cline{2-2} 
\multicolumn{1}{c|}{}                                                                                           & Only textual erasing& 79.21     & 79.55   \\ \cline{2-2} 
\multicolumn{1}{c|}{}                                                                                           & Only visual erasing & 79.05     & 79.37   \\ \hline
\multicolumn{2}{c|}{Iterative erasing}                                                                           & 80.13     & 79.97   \\ \hline
\multicolumn{2}{c|}{Erase during inference}                                                                                        & 79.25     & 79.56   \\ \hline
\multicolumn{2}{c|}{Multiple steps of attention}                                                                             & 79.31     & 78.49   \\ \hline
\end{tabular}
\caption{Ablation study results on RefCOCOg dataset.}
\label{tab:abliation}
\vspace{-10pt}
\end{table}
\noindent\textbf{Erasing methods.}
Different choices of erasing methods were exploited by previous works. Other than our proposed attention-guided erasing, the most straightforward way is to randomly erase words or image regions without considering their importances~\cite{singh2017hide}. Another choice is to train an adversarial network to select the most informative word or region to erase, which is used in~\cite{wang2017fast}. We compare our attention-guided erasing approach with those methods, and results in Table~\ref{tab:abliation} show that the attention-guided erasing performs better. Since attention weights are already good indicators of feature importance, leveraging attention as a guidance for erasing is more efficient, and the attention-guided erasing approach leads to little cost in model complexity, compared with applying a separate adversarial erasing network.

\vspace{1pt}
\noindent\textbf{Effect of cross-modal erasing.}
We compare our cross-modal erasing approach with erasing based on self-attention weights, where we only utilize information within the same modality for generating attention weights and performing attention-guided erasing. We also experiment on only visual erasing or sentence erasing. Experimental results in Table~\ref{tab:abliation} demonstrate the necessity of both visual erasing and query sentence erasing which are complementary to each other, and validate that our cross-modal attention-guided erasing is superior to self-attention-guided erasing without considering information from the other modality.

\vspace{1pt}
\noindent\textbf{Iterative erasing.}
A possible extension is to iteratively perform multiple times of erasing similar to~\cite{wei2017object} to generate more challenging training samples progressively. However, results in Table~\ref{tab:abliation} indicate that it is not suitable for this task. We observe that most referring expressions are quite short.
Erasing more than one key words would significantly eliminate the semantic meaning of the sentence. Likewise, erasing the visual features for more than once would also make it impossible for the model to recognize the referred object.

\vspace{1pt}
\noindent\textbf{Erasing during inference.}
Our model only leverages cross-modal erasing in the training phase and does not erase during inference. We try to erase key words or key regions during inference as well, and ensemble the matching scores of original samples and erased samples as the final score. But experiments suggest that it does not help the final performance. This is possibly because during training, the model have already learned to balance the weights of various features, and do not need to mask the dominant features to discover other alignments during inference.

\vspace{1pt}
\noindent\textbf{Comparison with stacked attention.}
Leveraging multiple steps of attention also enables the model to attend to different features. However, those models do not pose direct constraints on learning complementary attention for different attention steps. We conduct experiments on stacked attention~\cite{yang2016stacked} to compare with our erasing approach. Experiments indicate that erasing performs better than stacked attention on this task, because by erasing we enforce stricter constraints of learning complementary alignments.

\vspace{-3pt}
\section{Conclusion and Future Work} 


We address the problem of comprehending and aligning various types of information for referring expression grounding. To prevent the model from over-concentrating on the most significant cues and drive the model to discover complementary textual-visual alignments, we design a cross-modal attention-guided erasing approach to generate hard training samples by discarding the most important information. The models achieve state-of-the-art performance on three referring expression grounding datasets, demonstrating the effectiveness of our approach. 

\section*{Acknowledgements}

This work is supported in part by SenseTime Group Limited, in part by the General Research Fund through the Research Grants Council of Hong Kong under Grants CUHK14202217, CUHK14203118, CUHK14205615, CUHK14207814, CUHK14213616, CUHK14208417, CUHK14239816, and in part by CUHK Direct Grant. 

{\small
\bibliographystyle{ieee_fullname}
\bibliography{bib}
}

\end{document}